\documentclass[english]{article}
\usepackage[T1]{fontenc}
\usepackage[utf8]{inputenc}
\usepackage{geometry}
\usepackage{color}
\usepackage{booktabs}
\usepackage{units}
\usepackage{amsbsy}
\usepackage{amstext}
\usepackage{amssymb}
\usepackage{graphicx}

\makeatletter

\providecommand{\tabularnewline}{\\}

\@ifundefined{date}{}{\date{}}
\usepackage{tikz}
\usepackage{url}
\usepackage[sort,compress,space]{cite}

\@ifundefined{showcaptionsetup}{}{%
 \PassOptionsToPackage{caption=false}{subfig}}
\usepackage{subfig}
\makeatother

\usepackage{babel}
\begin{document}

\title{Modeling with Copulas and Vines in \\
Estimation of Distribution Algorithms}

\author{Marta Soto \\ \small{\vspace{-0.3em} Institute of Cybernetics, Mathematics and Physics, Cuba.} \\ \small{\vspace{-0.3em}Email: \texttt{mrosa@icimaf.cu}\vspace{0.6em}}\and Yasser González-Fernández \\ \small{\vspace{-0.3em} Institute of Cybernetics, Mathematics and Physics, Cuba.} \\ \small{\vspace{-0.3em}Email: \texttt{ygf@icimaf.cu}\vspace{0.6em}}\and Alberto Ochoa \\ \small{\vspace{-0.3em} Institute of Cybernetics, Mathematics and Physics, Cuba.} \\ \small{\vspace{-0.3em}Email: \texttt{ochoa@icimaf.cu}\vspace{0.6em}}}
\maketitle
\begin{abstract}
The aim of this work is studying the use of copulas and vines in the
optimization with Estimation of Distribution Algorithms (EDAs). Two
EDAs are built around the multivariate product and normal copulas,
and other two are based on pair-copula decomposition of vine models.
Empirically we study the effect of both marginal distributions and
dependence structure separately, and show that both aspects play a
crucial role in the success of the optimization. The results show
that the use of copulas and vines opens new opportunities to a more
appropriate modeling of search distributions in EDAs. 
\end{abstract}

\section{Introduction}

Estimation of distribution algorithms (EDAs) \cite{Larranaga2002EDANewToolEC,Muhlenbein1996BinaryParameters}
are stochastic optimization methods characterized by the explicit
use of probabilistic models. EDAs explore the search space by sampling
a probability distribution (search distribution) previously built
from promising solutions. 

Most existing continuous EDAs are based on either the multivariate
normal distribution or  models derived from it \cite{Bosman2006RealValuedEDAs,Kern2003ReviewContinuousEDAs}.
However,  in situations where empirical evidence reveals significant
departures from the normality assumption, these EDAs construct incorrect
models of the search space. A solution come with the copula function
\cite{Nelsen2006IntroductionCopulas}, which provides a way to separate
the statistical properties of each variable from the dependence structure:
first, the marginal distributions are fitted using a rich variety
of univariate models available, and then, the dependence between the
variables is modeled using a copula. However, the multivariate copula
approach has limitations. The number of multivariate copulas is rather
limited, and usually these copulas have only one parameter to describe
the overall dependence. Thus, this approach is not appropriate when
all the pairs of variables do not have the same type or strength of
dependence. For instance, the $t$-copula uses one correlation coefficient
per each pair of variables, but has only a single degree of freedom
parameter to characterize the tail dependence for all pairs.

An alternative approach to this problem is the pair-copula construction
method (PCC) \cite{Bedford2001DensityDecomposition,Bedford2002VinesNewGraphicalModel,Joe1996hFunctions},
which allows to built multivariate distributions using only bivariate
copulas. PCC models of multivariate distributions are represented
in a graphical way as a sequence of nested trees, which are called
vines. These graphical models provide a powerful and flexible tool
to deal with complex dependences as far as the pair-copulas in the
decomposition can be of different copula families.

In recent years, several copula-based EDAs have been proposed in the
literature. The authors have studied the behavior of these algorithms
in test functions \cite{Cuesta-Infante2010ArchimedeanCopulasEDA,Gao2009EDALaplaceTransform,Salinas-Gutierrez2009CopulasEDAs,Soto2007EDACopulaGaussiana,Wang2009EDACopula,Soto2010VEDA,Gonzalez-Fernandez2011BachelorThesis,Gonzalez-Fernandez2012MAEB}
and a real-world problem \cite{Soto2012VEDAMD}. Indeed, the use of
copulas has been identified as one of the emerging trends in the optimization
of real-valued problems using EDAs \cite{Hauschild2011EDASurveyPaper}.
In this work, various models based on copula theory are combined in
an EDA: two models are built using the multivariate product and normal
copulas and other two are based on two PCC models called C-vine and
D-vine. We empirically evaluate the performance of these algorithms
on a set of test functions and show that vine-based EDAs are better
endowed to deal with problems with different dependences between pair
of variables.

The paper is organized as follows. Section~\ref{sec:edas-copulas}
introduces the notion of copula and describes two EDAs based on the
multivariate product and normal copulas, respectively. Section~\ref{sec:edas-vines}
presents the notion and terminology of vines and introduces two EDAs
based on C-vine and D-vine models, respectively. Section~\ref{sec:experiments}
reports and discuses the empirical investigation. Finally, Section~\ref{sec:conclusions}
gives the conclusions.

\section{Two Continuous EDAs Based on Multivariate Copulas\label{sec:edas-copulas}}

We start with some definitions from copula theory \cite{Joe1997MultivariateModels,Nelsen2006IntroductionCopulas}.
Consider $n$ random variables $\mathbf{X}=\left(X_{1},\ldots,X_{n}\right)$
with joint cumulative distribution function $F$ and joint density
function $f$. Let $\mathbf{x}=\left(x_{1},\ldots,x_{n}\right)$ be
an observation of $\mathbf{X}$. A copula $C$ is a multivariate distribution
with uniformly distributed marginals $U\left(0,1\right)$ on $\left[0,1\right]$.
Sklar’s theorem \cite{Sklar1959CopulasFrench} states that every multivariate
distribution $F$ with marginals $F_{1},F_{2},\ldots,F_{n}$ can be
written as 
\[
F\left(x_{1},\ldots,x_{n}\right)=C\left(F\left(x_{1}\right),\ldots,F\left(x_{n}\right)\right)
\]

\noindent and

\[
C\left(u_{1},\ldots,u_{n}\right)=F\left(F_{1}^{(-1)}\left(u_{1}\right),\ldots,F_{n}^{(-1)}\left(u_{n}\right)\right)
\]
where $F_{i}^{(-1)}$ are the inverse distribution functions of the
marginals. If $F$ is continuous then $C\left(u_{1},\ldots,u_{n}\right)$
is unique. \textcolor{black}{The notion of copulas separates the effect
of dependence and margins in a joint distribution \cite{Kurowicka2006UncertaintyAnalysis}.
}The copula $C$ provides all information about the dependence structure
of $F$, independently of the specification of the marginal distributions.

An immediate consequence of Sklar’s theorem is that random variables
are independent if and only if their underlying copula is the independence
or product copula $C_{\textrm{I}}$, which is given by 

\begin{equation}
C_{\textrm{I}}\left(u_{1},\ldots,u_{n}\right)=u_{1}.\ldots.u_{n}.\label{eq:mv-product-copula}
\end{equation}

The UMDA proposed in \textcolor{black}{\cite{Larranaga2002EDANewToolEC}}
assumes a model of independence of normal margins. Therefore, an EDA
based on the product copula is a generalization of the UMDA, which
also supports other types of marginal distributions.

\textcolor{black}{Besides UMDA, in \cite{Larranaga2002EDANewToolEC}
the authors also proposed an EDA based on the multivariate normal
distribution. They called it Estimation of the Multivariate Normal
Algorithm (EMNA). It turns out that, indeed EMNA can be also reformulated
in copula terms: a normal copula plus normal margins.}

\textcolor{black}{The Gaussian Copula Estimation of Distribution Algorithm
(GCEDA) proposed in \cite{Soto2007EDACopulaGaussiana,Arderi2007BachelorThesis}
uses the multivariate normal (or Gaussian) copula, which is given
by}

\begin{equation}
C_{\textrm{N}}\left(u_{1},\ldots,u_{n};R\right)=\Phi_{R}\left(\Phi^{-1}\left(u_{1}\right),\ldots,\Phi^{-1}\left(u_{n}\right)\right),\label{eq:mv-normal-copula}
\end{equation}
\textcolor{black}{where $\Phi_{R}$ is the standard multivariate normal
distribution with correlation matrix $R$, and $\Phi^{-1}$ denotes
the inverse of the standard univariate normal distribution. This copula
allows the construction of multivariate distributions with non-normal
margins. If this is the case, the joint density is no longer the multivariate
normal, though the normal dependence structure is preserved. Therefore,
with normal margins, GCEDA is equal  to EMNA, otherwise they are
different.}

If the marginal distributions are non-normal, the correlation matrix
is estimated using the inversion of the non-parametric estimator Kendall's
tau $\hat{R}_{ij}=\sin\left(\nicefrac{\pi}{2}\hat{\tau}_{ij}\right)$
for each pair of variables $i,j=1,\ldots,n$ \cite{Nelsen2006IntroductionCopulas}.
If the resulting matrix $\hat{R}$ is not positive-definite, the correction
proposed in \cite{Rousseeuw1993TransformCorrMatrix} can be applied.

In this work, all margins used by the algorithms are always of the
same type, either normal (Gaussian) or empirical  smoothed with a
normal kernel. In particular, the estimation of the normal margin
$\hat{F_{i}}\backsim\mathcal{N}\left(x_{i};\hat{\mu}_{i},\hat{\sigma}_{i}^{2}\right)$
requires the computation of the mean $\hat{\mu_{i}}$ and variance
$\hat{\sigma_{i}}^{2}$ from the selected population. The empirical
margin is estimated using the normal kernel estimator given by 

\[
\hat{F}_{i}\left(t\right)=\frac{1}{N}\sum_{j=1}^{N}\Phi\left(\frac{t-y_{j}}{h}\right),
\]
where the set $\left\{ y_{1},\ldots,y_{N}\right\} $ is the sample
of the $i^{th}$ variable of $\mathbf{X}$ in the selected population
with $N$ individuals. The bandwidth parameter $h$ is computed according
to the rule-of-thumb of \cite{Silverman1986DensityEstimation}. In
this paper, the subscripts g and e in the name of the algorithms denote
the use of Gaussian and empirical margins, respectively (e.g., $\textrm{UMDA\ensuremath{{}_{\textrm{g}}}}$
and $\textrm{GCEDA\ensuremath{{}_{\textrm{e}}}}$). 

The generation of a new individual in $\textrm{GCEDA\ensuremath{{}_{\textrm{g}}}}$
and $\textrm{GCEDA\ensuremath{{}_{\textrm{e}}}}$ starts with the
simulation of a vector $\left(u_{1},\ldots,u_{n}\right)$ from the
multivariate normal copula \cite{Fusai2008StructuringDependence}.
In $\textrm{GCEDA\ensuremath{{}_{\textrm{g}}}}$, the inverse distribution
function $x_{i}=\hat{F}_{i}^{-1}\left(u_{i};\hat{\mu}_{i},\hat{\sigma}_{i}^{2}\right)$
is used to obtain each $x_{i}$ of the new individual. In $\textrm{GCEDA\ensuremath{{}_{\textrm{e}}}}$,
$x_{i}$ is found by solving the inverse of the marginal cumulative
distribution using the Newton-Raphson method \cite{Azzalini1981DistributionKernel}.

\section{EDAs Based on Vines\label{sec:edas-vines}}

This section provides a brief description of the C-vine and D-vine
models and the motivation for using them to construct the search distributions
in EDAs. We also introduce CVEDA and DVEDA, our third and fourth algorithms.

\subsection{From Multivariate Copulas to Vines}

The multivariate copula approach has several limitations. Most of
the available parametric copulas are bivariate and the multivariate
extensions usually describe the overall dependence by means of only
one parameter. This approach is not appropriate when there are pairs
of variables with different type or strength of dependence. The pair-copula
construction method (PCC) is an alternative approach to this problem.
PCC method was originally proposed in \cite{Joe1996hFunctions} and
this result was later developed in \cite{Bedford2001DensityDecomposition,Bedford2002VinesNewGraphicalModel,Joe1996hFunctions}.
The decomposition of a multivariate distribution in pair-copulas is
a general and flexible method for constructing multivariate distributions.
In PCC models, bivariate copulas are used as building blocks. The
graphical representation of these constructions involves a sequence
of nested trees, called regular vines. Pair-copula constructions of
regular vines allows to model a rich variety of types of dependences
as far as the bivariate copulas can belong to different families.

\subsection{Pair-Copula Constructions of C-vines and D-vines\label{sub:pcc-vines}}

Vines are dependence models of a multivariate distribution function
based on a decomposition of $f\left(x_{1},\ldots,x_{n}\right)$ into
bivariate copulas and marginal densities. A vine on $n$ variables
is a nested set of trees $T_{1},\ldots,T_{n-1}$, where the edges
of tree $j$ are the nodes of the tree $j+1$ with $j=1,\ldots,n-2$.
Regular vines constitute a special case of vines in which two edges
in tree $j$ are joined by an edge in tree $j+1$ only if these edges
share a common node. 

\begin{figure}[t]
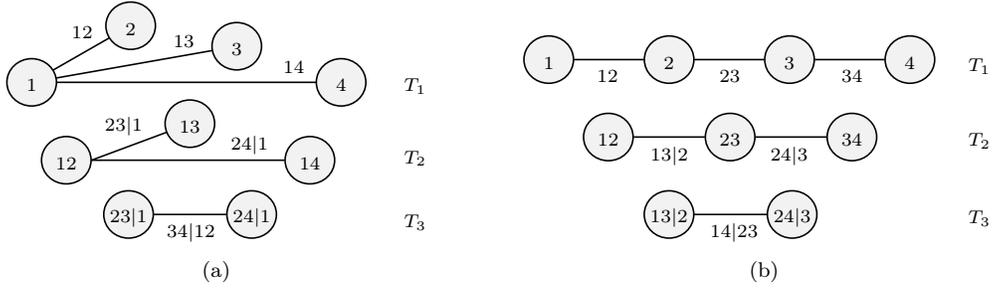

\begin{centering}
\subfloat[]{\scriptsize \input{C-vine-4d.tex}}\hspace{1cm}\subfloat[]{\scriptsize \input{D-vine-4d.tex}

}
\par\end{centering}

\caption{Four-dimensional C-vine (a) and D-vine (b). In a C-vine, each tree
$T_{j}$ has a unique node with $n-j$ edges. The node with $n-1$
edges in tree is called the root. In a D-vine, no node is connected
to more than two edges. \label{fig:c-vine-d-vine}}
\end{figure}

Two instances of regular vines are the canonical (C) and drawable
(D) vines. In Figure~\ref{fig:c-vine-d-vine}, a graphical representation
of a C-vine and D-vine for four dimensions is given. Each graphical
model gives a specific way of decomposing the density. In particular,
for a C-vine, $f\left(x_{1},\ldots,x_{n}\right)$ is given by 

\begin{equation}
\prod_{k=1}^{n}f\left(x_{k}\right)\prod_{j=1}^{n-1}\prod_{i=1}^{n-j}c_{j,,j+i|i,\ldots,,j-1}\left(F\left(x_{j}|x_{1},\ldots,x_{j\text{-}1}\right),F\left(x_{j+i}|x_{1},\ldots,x_{j-1}\right)\right),\label{eq:c-vine}
\end{equation}
and for a D-vine, the density is equal to

\begin{equation}
\prod_{k=1}^{n}f\left(x_{k}\right)\prod_{j=1}^{n-1}\prod_{i=1}^{n-j}c_{i,i+j|i+1,\ldots,i+j-1}\left(F\left(x_{i}|x_{i+1},\ldots,x_{i+j-1}\right),F\left(x_{i+j}|x_{i+1},\ldots,x_{i+j\text{-}1}\right)\right),\label{eq:d-vine}
\end{equation}
{\footnotesize{} }where $j$ identifies the trees and $i$ denotes
the edges in each tree.

Note that in (\ref{eq:c-vine}) and (\ref{eq:d-vine}) the joint density
consists of marginal densities $f\left(x_{k}\right)$ and pair-copula
densities evaluated at conditional distribution functions of the form
$F\left(x\mid\mathbf{v}\right)$. 

In \cite{Joe1996hFunctions} it is showed that conditional distribution
of pair-copulas constructions are given by

\begin{equation}
F\left(x\mid\mathbf{v}\right)=\frac{\partial C_{xv_{j}\mid\mathbf{v}_{-j}}\left(F\left(x\mid\mathbf{v}_{-j}\right),F\left(v_{j}\mid\mathbf{v}_{-j}\right)\right)}{\partial F\left(v_{j}\mid\mathbf{v}_{-j}\right)},\label{eq: Cond Dist}
\end{equation}
where $C_{xv_{j}\mid\mathbf{v}_{-j}}$ is a bivariate copula distribution
function, $\mathbf{v}$ is a $n$-dimensional vector, $v_{j}$ is
the $j$ components of $\mathbf{v}$ and $\mathbf{v}_{-j}$ denotes
the remaining component. The recursive evaluation of $F\left(x\mid\mathbf{v}\right)$
yields the expression 

\[
F\left(x\mid v\right)=\frac{\partial C_{xv}\left(F_{x}\left(x\right),F_{v}\left(v\right)\right)}{\partial F_{v}\left(v\right)}.
\]
For the special case (unconditional) when $v$ is univariate, and
$x$ and $v$ are standard uniform, $F\left(x\mid v\right)$ reduces
further to

\[
F\left(x\mid v\right)=\frac{\partial C_{xv}\left(x,v,\Theta\right)}{\partial v}.
\]
where $\mathbf{\Theta}$ is the set of parameters for the bivariate
copula of the joint distribution function of $x$ and $v$. To facilitate
de computation of $F\left(x\mid v\right)$, the function
\begin{equation}
h\left(x,v;\theta\right)=F\left(x\mid v\right)=\frac{\partial C_{xv}\left(x,v;\Theta\right)}{\partial v},\label{eq: h-function}
\end{equation}
is defined. The inverse of $h$ with respect to the first variable
$h^{-1}$ is also defined. The expressions of these functions of the
bivariate copulas used in this work are given in Appendix~\ref{sec:h-functions}.

\subsection{Vine Estimation of Distribution Algorithms}

Vine Estimation of Distribution Algorithms (VEDAs) \cite{Gonzalez-Fernandez2011BachelorThesis,Soto2010VEDA}
are a class of EDAs that uses vines to model the search distributions.
CVEDA and DVEDA are VEDAs based on C-vines and D-vines, respectively.
Now we describe the particularities of the estimation and simulation
steps of these algorithms.

\subsubsection{Estimation \label{sub: Estimation}}

The estimation procedures of C-vines and D-vines proposed and developed
in \cite{Aas2009PairCopulaConstructions}  consist of the following
main steps: selection of a specific factorization, choice of the pair-copula
types in the factorization, and estimation of the copula parameters.
Next we describe these steps according to our implementation. 
\begin{enumerate}
\item Selection of a specific factorization:

\textcolor{black}{The selection of a }specific\textcolor{black}{{} pair-copula
decomposition implies to choose an appropriate order of the variables,
which can be obtained by several ways: given as parameter, selected
at random, chosen by greedy heuristics. We use greedy heuristics for
detecting the most important bivariate dependences.}

\textcolor{black}{Assumed a specific factorization, the first step
of the estimation procedure consist in  assigning weights to the
edges. As weight we use the absolute value of the empirical Kendall's
tau between pair of variables }\cite{Nelsen2006IntroductionCopulas}\textcolor{black}{.
The next step consist in determining the appropriate order of the
variables of the decomposition, which depend on the type of pair-copula
decomposition: }
\begin{itemize}
\item \textcolor{black}{In a C-vine, the tree that maximizes the sum of
the weights of one node (the root) to the others is chosen by the
greedy heuristic as the appropriate factorization. }
\item \textcolor{black}{In a D-vine, the first tree is that which maximizes
the weighted sequence of the original variables. In \cite{Brechmann2010DiplomaThesis},
this problem is transformed into a traveling salesman problem (TSP)
instance by adding a dummy node with weight zero on all edges to the
other nodes. For efficiency, we use the cheapest insertion heuristic,
an approximate solution of TSP presented in \cite{Rosenkrantz1977TSPHeuristics}.
In a D-vine, the structure of remaining trees is completely determined
by the structure of the first.}
\end{itemize}

A pair-copula decomposition has $n-1$ trees and requires to fit $\nicefrac{n\left(n-1\right)}{2}$
copulas. Assuming conditional independence might simplify the estimation
step, since\textcolor{black}{{} if $X$ and $Y$ are conditionally independent
given $\mathbf{V}$, then $c_{xy\mid\mathbf{v}}\left(F_{x\mid\mathbf{v}}\left(x\mid\mathbf{v}\right),\, F_{y\mid\mathbf{v}}\left(y\mid\mathbf{v}\right)\right)=1$.
This property is used by a model selection procedure proposed in \cite{Brechmann2010DiplomaThesis},
which consists in truncating the pair-copula decomposition at specific
tree level, fitting the product copula in the subsequent trees. For
detecting the truncation tree level, this procedure uses either the
Akaike Information Criterion (AIC) \cite{Akaike1974AIC} or the Bayesian
Information Criterion (BIC) \cite{Schwarz1978BIC},} such that the
tree $T_{j+1}$ is expanded if the value of the information criteria
calculated up to the tree $T_{j+1}$ is smaller than the value obtained
up to the previous tree. Otherwise, the vine is truncated at tree
level $T_{j}$. 

\item Choice of the pair-copula types in the factorization and estimation
of the copula parameters.

\begin{enumerate}
\item \textcolor{black}{Determine which pair-copula types to use in tree
1 using the original data by applying a goodness of fit test. }
\item \textcolor{black}{Compute observations (i.e. conditional distribution
functions) using the copula parameters from tree 1 and the $h\left(.\right)$
function.}
\item \textcolor{black}{Determine the pair-copula types to use in tree 2
in the same way as in tree 1 using the }observations from (b\textcolor{black}{).}
\item Repeat (b) and \textcolor{black}{(c)} for the following trees. 
\end{enumerate}

Selection of pair-copulas is accomplished in different ways \cite{Genest2009GoFTestCopulas}.
 In this work, the Cramér-von Mises statistics 
\begin{equation}
S_{N}=\sum_{i=1}^{N}\left(C_{\textrm{\ensuremath{\mbox{E}}}}(u_{i,}v_{i})-C_{\mathbf{\Theta}}(u_{i},v_{i})\right)^{2}\label{eq: Cramer-von Mises statisic}
\end{equation}
is minimized. $N$ is the sample size, $\mathbf{\Theta}$ is the set
of parameters of a bivariate copula $C_{\mathbf{\Theta}}$, and $C_{\textrm{\ensuremath{\mbox{E}}}}$
is the empirical copula. We first test the product copula \textcolor{black}{\cite{Genest2004TestIndepEmpiricalCopula}}.
If there is enough evidence against the null hypothesis of independence
(at a fixed significance level of $0.1$) it is rejected. If this
is the case, the copula $C_{\mathbf{\Theta}}$ that minimizes $S_{N}$
is chosen.

We combine different types of bivariate copulas: normal, Student's
$t$, Clayton, rotated Clayton, Gumbel and rotated Gumbel. The normal
copula is neither lower nor upper tail dependent while the Student’s
$t$ copula is both lower and upper tail dependent. The Clayton and
rotated Clayton copulas are lower tail dependent while the Gumbel
and rotated Gumbel copulas are upper tail dependent.

The parameters of all these copulas, but the Student's $t$, are estimated
using the inversion of Kendall's tau \cite{Genest2007CopulaAfraidAsk}.
The correlation coefficient for the Student's and normal copulas are
computed similarly. The degrees of freedom of the Student's $t$ copula
are estimated by maximum likelihood with the correlation parameter
held fixed \cite{Demarta2005tCopula}. We consider an upper bound
of $30$ for the degrees of freedom because for this value the bivariate
Student's $t$ copula becomes almost indistinguishable from the bivariate
normal copula \cite{Fantazzini2010EstimationTCopulas}.

\end{enumerate}

\subsubsection{Simulation}

\textcolor{black}{Simulation from vines \cite{Bedford2001MonteCarloSimulation,Bedford2001DensityDecomposition,Kurowicka2007SamplingVines}
is based on the conditional distribution method described in \cite{Devroye1986RandomVariateGeneration}.
The general algorithm for sampling $n$ dependent uniform $\left[0,1\right]$
variables is common for C-vines and D-vines. First, sample $n$ independent
uniform random numbers $w_{i}\in\left(0,1\right)$ and then compute}

\begin{center}
$\begin{array}{lll}
x_{1} & = & w_{1}\\
x_{2} & = & F_{2\mid1}^{-1}\left(w_{2}|x_{1}\right)\\
x_{3} & = & F_{3\mid1,2}^{-1}\left(w_{3}|x_{1},x_{2}\right)\\
\;\vdots\\
x_{n} & = & F_{n\mid1,2,\ldots,n-1}^{-1}\left(w_{n}|x_{1},\ldots,x_{n-1}\right).
\end{array}$
\par\end{center}

\textcolor{black}{To determine $F\left(x_{j}\mid x_{1},x_{2},\ldots,x_{j-1}\right)$
for each $j$, the expressions (\ref{eq: Cond Dist}) and (\ref{eq: h-function})
are used for both structures, although the choice of the $v_{j}$
in (\ref{eq: Cond Dist}) is different (see (\ref{eq:c-vine}) and
(\ref{eq:d-vine})). For details about the sampling algorithms, see
\cite{Aas2009PairCopulaConstructions}.}

\section{Empirical Investigation\label{sec:experiments}}

This section outlines the experimental setup and presents the numerical
results. The experiments aim to show that both aspects,  the marginal
distributions and the dependence structure, are crucial for EDA optimization. 

For the empirical study we use the statistical environment R \cite{RDevelopmentCoreTeam2011R}
and the tools provided by the packages\texttt{ copulaedas} \cite{Gonzalez-Fernandez2011copulaedasRPackage}
and \texttt{vines} \cite{Gonzalez-Fernandez2011vinesRPackage}.

\subsection{Experimental Design}

The well known Sphere, Griewank, Ackley and Summation Cancellation
test functions \cite{Bengoetxea2002ExperimentalResultsEDAs} are considered
as benchmark problems in $n=10$ dimensions. The definition of these
functions for $\boldsymbol{x}=(x_{1},\ldots,x_{n})$ is given below:

\vspace{-0.5em}

\[
f_{\textrm{Sphere}}(\boldsymbol{x})=\sum_{i=1}^{n}x_{i}^{2}
\]

\vspace{-1em}

\[
f_{\textrm{Griewank}}(\boldsymbol{x})=1+\sum_{i=1}^{n}\frac{x_{i}^{2}}{4000}-\prod_{i=1}^{n}\cos\left(\frac{x_{i}}{\sqrt{i}}\right)
\]

\vspace{-1em}

\[
f_{\textrm{Ackley}}(\boldsymbol{x})=-20\exp\left(-0.2\sqrt{\frac{1}{n}\sum_{i=1}^{n}x^{2}}\right)-\exp\left(\frac{1}{n}\sum_{i=1}^{n}\cos\left(2\pi x_{i}\right)\right)+20+\exp\left(1\right)
\]

\vspace{-1em}

\[
f_{\textrm{Summation\ensuremath{\,}Cancellation}}(\boldsymbol{x})=\frac{1}{10^{-5}+\sum_{i=1}^{n}|y_{i}|},\, y_{1}=x_{1},\, y_{i}=y_{i-1}+x_{i}
\]

Sphere, Griewank and Ackley are minimization problems that have global
optimum at $\boldsymbol{x}=(0,\ldots,0)$ with evaluation zero. Summation
Cancellation is a maximization problem that has global optimum at
$\boldsymbol{x}=(0,\ldots,0)$ with evaluation $10^{5}$.

To ensure a fair comparison between the algorithms, we find the minimum
population size required by each algorithm to reach the global optimum
of the function in $30$ of $30$ independent runs. This critical
population size is determined using a bisection method \cite{Pelikan2005hBOA}.
The algorithm  stops when either the global optimum is found with
a precision of $10^{-6}$ or after $500,000$ function evaluations.
A truncation selection of $0.3$ is used \cite{Muhlenbein1999FDA},
and no elitism. 

In the initial population, each variable is sampled uniformly in a
given real interval. We say an interval is symmetric if the value
that $X_{i}$ takes in the global optimum of the function is located
in the middle of the given interval. Otherwise, we call it asymmetric.
The symmetric intervals used in the experiments are: $[-600,600]$
in Sphere and Griewank, $[-30,30]$ in Ackley, and $[-0.16,\,0.16]$
in Summation Cancellation. The asymmetric intervals are: $[-300,900]$
in Sphere and Griewank, $[-15,45]$ in Ackley, and $[-0.08,\,0.24]$
in Summation Cancellation.

\subsection{Effect of the Marginal Distributions \label{sub: Effec of Marginal Distributions}}

In this section we investigate the effect of the marginal distributions
under two assumptions: independence and joint normal dependence. The
results obtained with UMDA and GCEDA in symmetric and asymmetric intervals
are given in Tables \ref{tab:sphere-umda-gceda}--\ref{tab:sumcan-umda-gceda}.
We summarize the results in the following four points.

\begin{table}[t]
\caption{Results of UMDA and GCEDA in Sphere. \label{tab:sphere-umda-gceda}}

\centering{}%
\begin{tabular}{ccccc}
\toprule 
{\small Algorithm} & {\small Success } & {\small Population} & {\small Evaluations} & {\small Best Evaluation}\tabularnewline
\midrule
\addlinespace[0.4em]
\multicolumn{5}{c}{{\small $X_{i}\in[-600,600],\, i=1,\ldots,10$}}\tabularnewline\addlinespace[0.15em]
{\small $\textrm{UMDA\ensuremath{{}_{\textrm{g}}}}$} & {\small $30/30$} & {\small $86$} & {\small $3,996.1\mathbf{\pm}89.5$} & {\small $6.9\textrm{E}-07\pm1.9\textrm{E}-07$}\tabularnewline
{\small $\textrm{UMDA\ensuremath{{}_{\textrm{e}}}}$} & {\small $30/30$} & {\small $82$} & {\small $5,466.6\pm164.4$} & {\small $7.0\textrm{E}-07\pm1.7\textrm{E}-07$}\tabularnewline
{\small $\textrm{GCEDA\ensuremath{{}_{\textrm{g}}}}$} & {\small $30/30$} & {\small $325$} & {\small $13,769.1\mathbf{\pm}248.5$} & {\small $6.6\textrm{E}-07\pm1.6\textrm{E}-07$}\tabularnewline
{\small $\textrm{GCEDA\ensuremath{{}_{\textrm{e}}}}$} & {\small $30/30$} & {\small $259$} & {\small $14,581.7\pm403.2$} & {\small $7.1\textrm{E}-07\pm2.0\textrm{E}-07$}\tabularnewline
\addlinespace[0.25em]
\multicolumn{5}{c}{{\small $X_{i}\in[-300,900],\, i=1,\ldots,10$}}\tabularnewline\addlinespace[0.15em]
{\small $\textrm{UMDA\ensuremath{{}_{\textrm{g}}}}$} & {\small $30/30$} & {\small $118$} & {\small $5,502.7\mathbf{\pm}125.8$} & {\small $6.4\textrm{E}-07\pm1.9\textrm{E}-07$}\tabularnewline
{\small $\textrm{UMDA\ensuremath{{}_{\textrm{e}}}}$} & {\small $30/30$} & {\small $83$} & {\small $5,513.9\pm180.6$} & {\small $7.4\textrm{E}-07\pm1.9\textrm{E}-07$}\tabularnewline
{\small $\textrm{GCEDA\ensuremath{{}_{\textrm{g}}}}$} & {\small $24/30$} & {\small $2000$} & {\small $171,666.6\pm166,976.4$} & {\small $3.0\textrm{E}+01\pm8.4\textrm{E}+01$}\tabularnewline
{\small $\textrm{GCEDA\ensuremath{{}_{\textrm{e}}}}$} & {\small $30/30$} & {\small $522$} & {\small $29,023.2\mathbf{\pm}541.4$} & {\small $7.2\textrm{E}-07\pm2.3\textrm{E}-07$}\tabularnewline
\bottomrule
\end{tabular}
\end{table}
 
\begin{table}[t]
\caption{Results of UMDA and GCEDA in Griewank. \label{tab:griewank-umda-gceda}}

\centering{}%
\begin{tabular}{lcccc}
\toprule 
{\small Algorithm} & {\small Success} & {\small Population} & {\small Evaluations} & {\small Best Evaluation}\tabularnewline
\midrule
\addlinespace[0.4em]
\multicolumn{5}{c}{{\small $X_{i}\in[-600,600],\, i=1,\ldots,10$}}\tabularnewline\addlinespace[0.15em]
{\small $\textrm{UMDA\ensuremath{{}_{\textrm{g}}}}$} & {\small $30/30$} & {\small $113$} & {\small $5,179.1\mathbf{\pm}210.0$} & {\small $7.2\textrm{E}-07\pm1.7\textrm{E}-07$}\tabularnewline
{\small $\textrm{UMDA\ensuremath{{}_{\textrm{e}}}}$} & {\small $30/30$} & {\small $475$} & {\small $27,961.6\pm1,387.5$} & {\small $7.0\textrm{E}-07\pm1.8\textrm{E}-07$}\tabularnewline
{\small $\textrm{GCEDA\ensuremath{{}_{\textrm{g}}}}$} & {\small $30/30$} & {\small $304$} & {\small $12,798.4\mathbf{\pm}351.1$} & {\small $6.6\textrm{E}-07\pm1.7\textrm{E}-07$}\tabularnewline
{\small $\textrm{GCEDA\ensuremath{{}_{\textrm{e}}}}$} & {\small $30/30$} & {\small $324$} & {\small $17,895.6\pm536.0$} & {\small $6.7\textrm{E}-07\pm1.7\textrm{E}-07$}\tabularnewline
\addlinespace[0.25em]
\multicolumn{5}{c}{{\small $X{}_{i}\in[-300,900],\, i=1,\ldots,10$}}\tabularnewline\addlinespace[0.15em]
{\small $\textrm{UMDA\ensuremath{{}_{\textrm{g}}}}$} & {\small $30/30$} & {\small $110$} & {\small $5,261.6\mathbf{\pm}284.6$} & {\small $6.7\textrm{E}-07\pm2.1\textrm{E}-07$}\tabularnewline
{\small $\textrm{UMDA\ensuremath{{}_{\textrm{e}}}}$} & {\small $30/30$} & {\small $449$} & {\small $26,580.8\pm1,003.3$} & {\small $7.3\textrm{E}-07\pm1.7\textrm{E}-07$}\tabularnewline
{\small $\textrm{GCEDA\ensuremath{{}_{\textrm{g}}}}$} & {\small $22/30$} & {\small $2000$} & {\small $201,333.3\pm183,220.5$} & {\small $1.3\textrm{E}-01\pm2.5\textrm{E}-01$}\tabularnewline
{\small $\textrm{GCEDA\ensuremath{{}_{\textrm{e}}}}$} & {\small $30/30$} & {\small $588$} & {\small $32,438.0\mathbf{\pm}860.9$} & {\small $8.0\textrm{E}-07\pm1.5\textrm{E}-07$}\tabularnewline
\bottomrule
\end{tabular}
\end{table}
 
\begin{table}[t]
\caption{Results of UMDA and GCEDA in Ackley. \label{tab:ackley-umda-gceda}}

\centering{}%
\begin{tabular}{ccccc}
\toprule 
{\small Algorithm} & {\small Success} & {\small Population} & {\small Evaluations} & {\small Best Evaluation}\tabularnewline
\midrule
\addlinespace[0.4em]
\multicolumn{5}{c}{{\small $X_{i}\in[-30,30],\, i=1,\ldots,10$}}\tabularnewline\addlinespace[0.15em]
{\small $\textrm{UMDA\ensuremath{{}_{\textrm{g}}}}$} & {\small $30/30$} & {\small $88$} & {\small $5,426.6\mathbf{\pm}127.2$} & {\small $8.2\textrm{E}-07\pm1.0\textrm{E}-07$}\tabularnewline
{\small $\textrm{UMDA\ensuremath{{}_{\textrm{e}}}}$} & {\small $30/30$} & {\small $94$} & {\small $8,024.4\pm210.1$} & {\small $8.6\textrm{E}-07\pm8.4\textrm{E}-08$}\tabularnewline
{\small $\textrm{GCEDA\ensuremath{{}_{\textrm{g}}}}$} & {\small $30/30$} & {\small $325$} & {\small $18,178.3\mathbf{\pm}207.8$} & {\small $8.0\textrm{E}-07\pm1.5\textrm{E}-07$}\tabularnewline
{\small $\textrm{GCEDA\ensuremath{{}_{\textrm{e}}}}$} & {\small $30/30$} & {\small $303$} & {\small $21,866.5\pm338.3$} & {\small $8.1\textrm{E}-07\pm1.4\textrm{E}-07$}\tabularnewline
\addlinespace[0.25em]
\multicolumn{5}{c}{{\small $X_{i}\in[-15,45],\, i=1,\ldots,10$}}\tabularnewline\addlinespace[0.15em]
{\small $\textrm{UMDA\ensuremath{{}_{\textrm{g}}}}$} & {\small $30/30$} & {\small $95$} & {\small $5,959.6\mathbf{\pm}111.3$} & {\small $7.7\textrm{E}-07\pm1.1\textrm{E}-07$}\tabularnewline
{\small $\textrm{UMDA\ensuremath{{}_{\textrm{e}}}}$} & {\small $30/30$} & {\small $91$} & {\small $7,995.8\pm183.1$} & {\small $8.3\textrm{E}-07\pm1.1\textrm{E}-07$}\tabularnewline
{\small $\textrm{GCEDA\ensuremath{{}_{\textrm{g}}}}$} & {\small $30/30$} & {\small $782$} & {\small $45,460.2\pm532.8$} & {\small $8.0\textrm{E}-07\pm1.2\textrm{E}-07$}\tabularnewline
{\small $\textrm{GCEDA\ensuremath{{}_{\textrm{e}}}}$} & {\small $30/30$} & {\small $357$} & {\small $26,013.4\mathbf{\pm}493.7$} & {\small $8.5\textrm{E}-07\pm8.2\textrm{E}-08$}\tabularnewline
\bottomrule
\end{tabular}
\end{table}
 
\begin{table}[t]
\caption{Results of UMDA and GCEDA in Summation Cancellation. \label{tab:sumcan-umda-gceda}}

\centering{}%
\begin{tabular}{ccccc}
\toprule 
{\small Algorithm} & {\small Success} & {\small Population} & {\small Evaluations} & {\small Best Evaluation}\tabularnewline
\midrule
\addlinespace[0.4em]
\multicolumn{5}{c}{{\small $X_{i}\in[-0,16,\,0,16],\, i=1,\ldots,10$}}\tabularnewline\addlinespace[0.15em]
{\small $\textrm{UMDA\ensuremath{{}_{\textrm{g}}}}$} & {\small $0/30$} & {\small $2000$} & {\small $500,000.0\pm0,0$} & {\small $6.9\textrm{E}+02\pm5.0\textrm{E}+02$}\tabularnewline
{\small $\textrm{UMDA\ensuremath{{}_{\textrm{e}}}}$} & {\small $0/30$} & {\small $2000$} & {\small $500,000.0\pm0,0$} & {\small $1.0\textrm{E}+03\pm1.2\textrm{E}+03$}\tabularnewline
{\small $\textrm{GCEDA\ensuremath{{}_{\textrm{g}}}}$} & {\small $30/30$} & {\small $325$} & {\small $38,848.3\mathbf{\pm}327,6$} & {\small $1.0\textrm{E}+05\pm1.2\textrm{E}-07$}\tabularnewline
{\small $\textrm{GCEDA\ensuremath{{}_{\textrm{e}}}}$} & {\small $30/30$} & {\small $1525$} & {\small $213,144.1\pm1,907.3$} & {\small $1.0\textrm{E}+05\pm1.0\textrm{E}-07$}\tabularnewline
\addlinespace[0.25em]
\multicolumn{5}{c}{{\small $X_{i}\in[-0,08,\,0,24],\, i=1,\ldots,10$}}\tabularnewline\addlinespace[0.15em]
{\small $\textrm{UMDA\ensuremath{{}_{\textrm{g}}}}$} & {\small $0/30$} & {\small $2000$} & {\small $500,000.0\pm0.0$} & {\small $5.6\textrm{E}+02\pm3.8\textrm{E}+02$}\tabularnewline
{\small $\textrm{UMDA\ensuremath{{}_{\textrm{e}}}}$} & {\small $0/30$} & {\small $2000$} & {\small $500,000.0\pm0.0$} & {\small $1.9\textrm{E}+03\pm1.9\textrm{E}+03$}\tabularnewline
{\small $\textrm{GCEDA\ensuremath{{}_{\textrm{g}}}}$} & {\small $4/30$} & {\small $2000$} & {\small $467,000.0\pm85,577.5$} & {\small $1.3\textrm{E}+04\pm3.4\textrm{E}+04$}\tabularnewline
{\small $\textrm{GCEDA\ensuremath{{}_{\textrm{e}}}}$} & {\small $30/30$} & {\small $1525$} & {\small $215,330.0\mathbf{\pm}1,621.8$} & {\small $1.0\textrm{E}+05\pm1.1\textrm{E}-07$}\tabularnewline
\bottomrule
\end{tabular}
\end{table}

\begin{enumerate}
\item As the asymmetry of the interval grows the performance of all the
algorithms deteriorate. This effect is larger with normal margins.

We illustrate this point through the analysis of the UMDA behavior.
With symmetric intervals, $\textrm{UMDA\ensuremath{{}_{\textrm{g}}}}$
outperforms $\textrm{UMDA\ensuremath{{}_{e}}}$, which is particularly
notable in the Griewank function. As example, Figure~\ref{fig:griewank-boxplot}
illustrates that the variance of the normal margin shrinks faster
than the variance of the normal kernel margin. The larger variance
of the empirical margin can be explained by the existence of global
and local optima, all of which are captured by the normal kernel margins.
Figure~\ref{fig:griewank-normal-kernel}-(left) shows several peaks
located near the values that the variable takes in the global and
local optima, while in Figure~\ref{fig:griewank-normal-kernel}-(right)
the peak of the normal density lies in the middle of the interval
regardless of the shape of the data. For this same reason, with symmetric
interval, the algorithms behave better with normal margins than with
empirical.

\begin{figure}[t]
\begin{centering}
\includegraphics[width=0.8\textwidth]{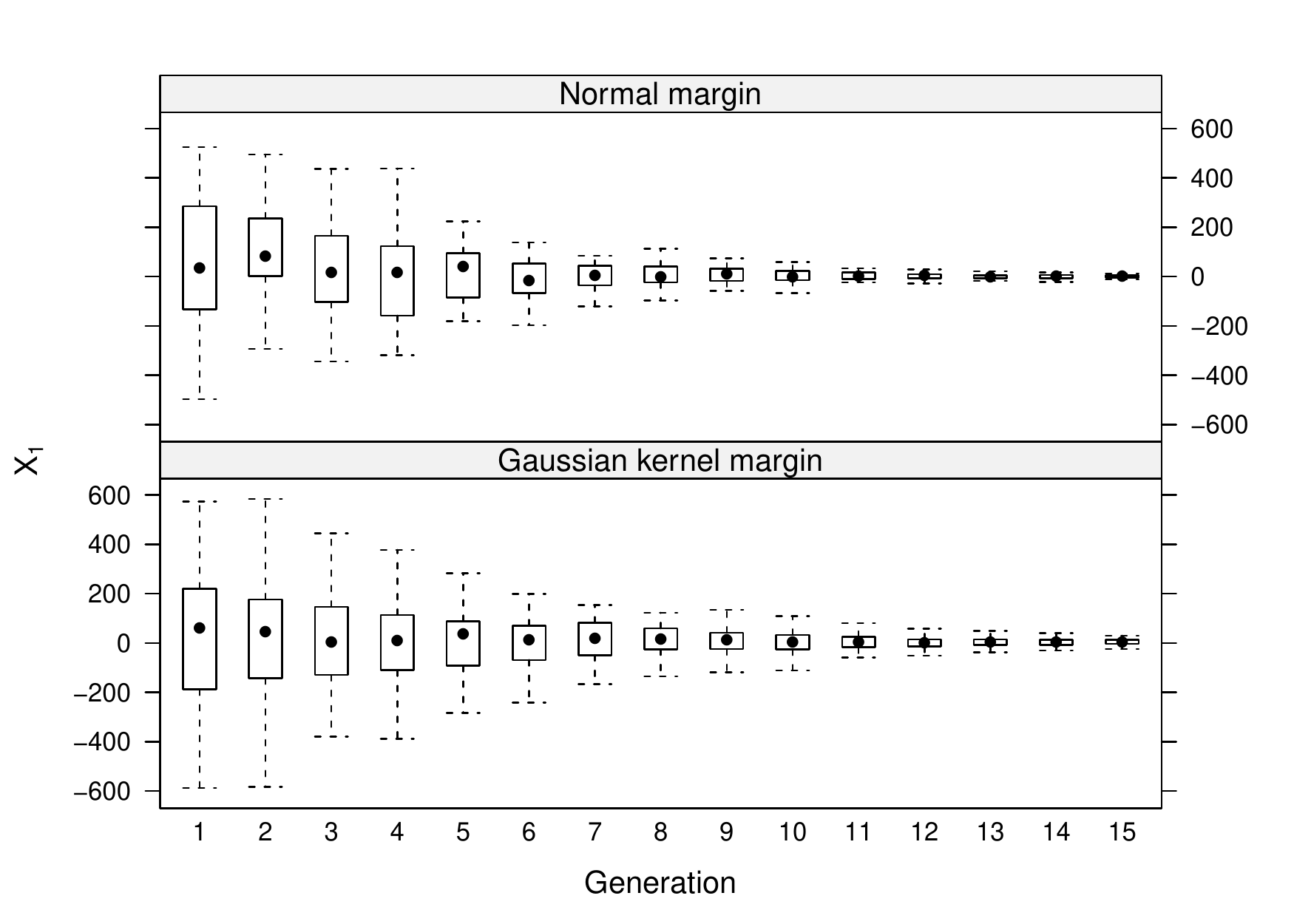}
\par\end{centering}

\caption{Box-plots illustrating the evolution of the first variable of Griewank
in the selected population of $\textrm{UMDA\ensuremath{{}_{\textrm{g}}}}$
(top) and $\textrm{UMDA\ensuremath{{}_{e}}}$ (bottom) for 15 generations.
\label{fig:griewank-boxplot}}
\end{figure}
\begin{figure}[t]
\begin{centering}
\includegraphics[width=0.75\textwidth]{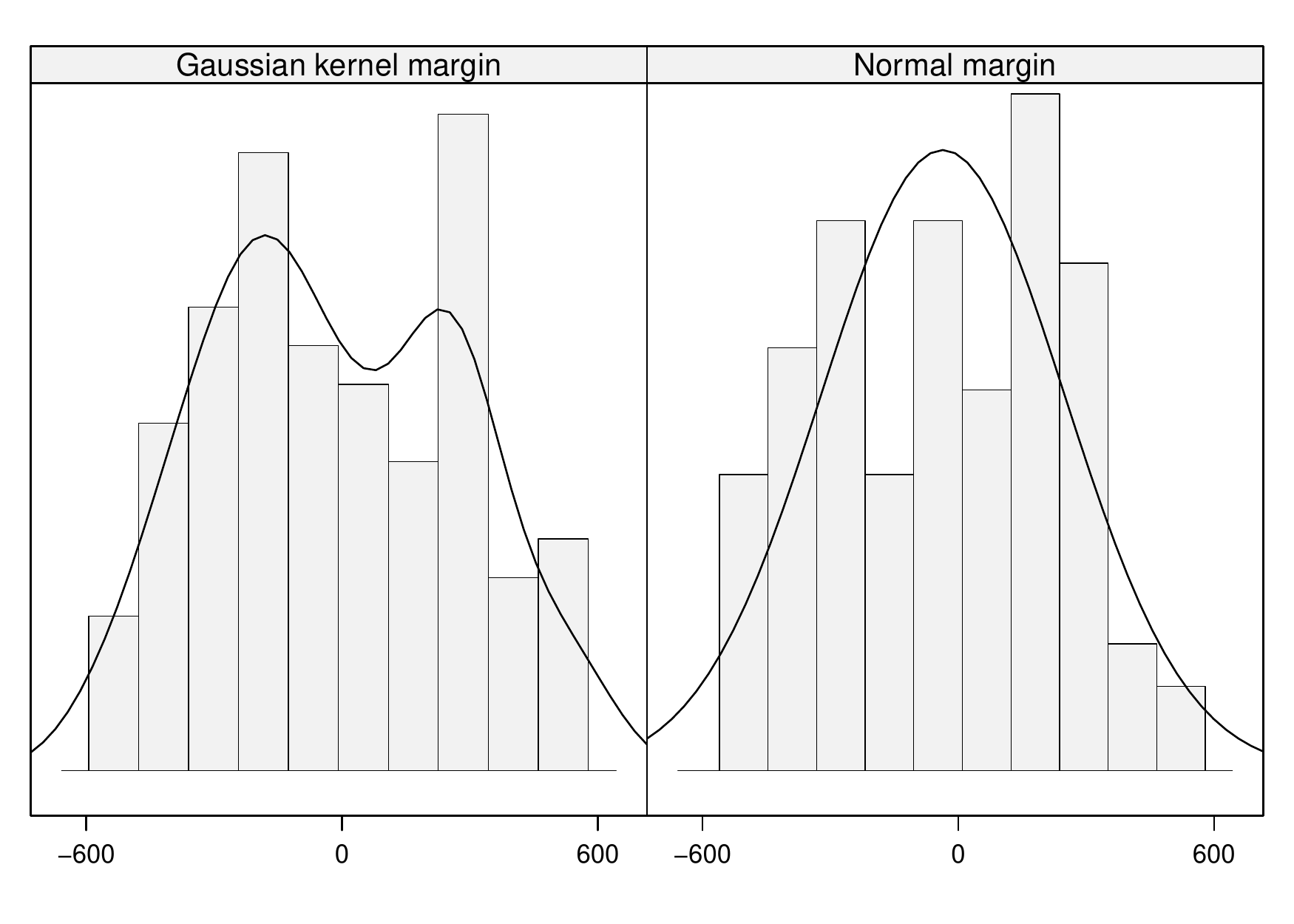}
\par\end{centering}

\caption{Histograms of the first variable of the Griewank function in the selected
population of the second generation with $\textrm{UMDA\ensuremath{{}_{\textrm{e}}}}$
(left) and $\textrm{UMDA\ensuremath{{}_{\textrm{g}}}}$ (right). The
empirical and normal densities are superposed, respectively. \label{fig:griewank-normal-kernel}}
\end{figure}

\item With asymmetric intervals, GCEDA with normal kernel margins is much
better than with normal margins.

With symmetric intervals, UMDA and GCEDA with normal margins behave
better than with normal kernel margins. However, if the initial population
is sampled asymmetrically, this situation changes, which is more remarkable
in GCEDA (even $\textrm{GCEDA\ensuremath{{}_{\textrm{g}}}}$ might
not converge). This situation is illustrated in the optimization of
the Griewank function with $\textrm{GCEDA\ensuremath{{}_{\textrm{g}}}}$
and $\textrm{GCEDA\ensuremath{{}_{\textrm{e}}}}$. Figure~\ref{fig:griewank-margins}
shows both the normal and normal kernel densities of the first variable,
which are estimated at generations $10$, $15$, $20$, $25$ and
$30$. We recall that the zero value corresponds to the value of the
variable in the global optimum. In Figure~\ref{fig:griewank-margins}-(top),
note that with normal margins the zero is located at  the tail of
the normal density, thus, it is sampled with low probability. As the
evolution proceeds, the density moves away from zero. In Figure~\ref{fig:griewank-margins}-(bottom),
the normal kernel margins capture more local features of the distribution
and it is more likely that good points are sampled.

\begin{figure}[t]
\begin{centering}
\includegraphics[width=0.8\textwidth]{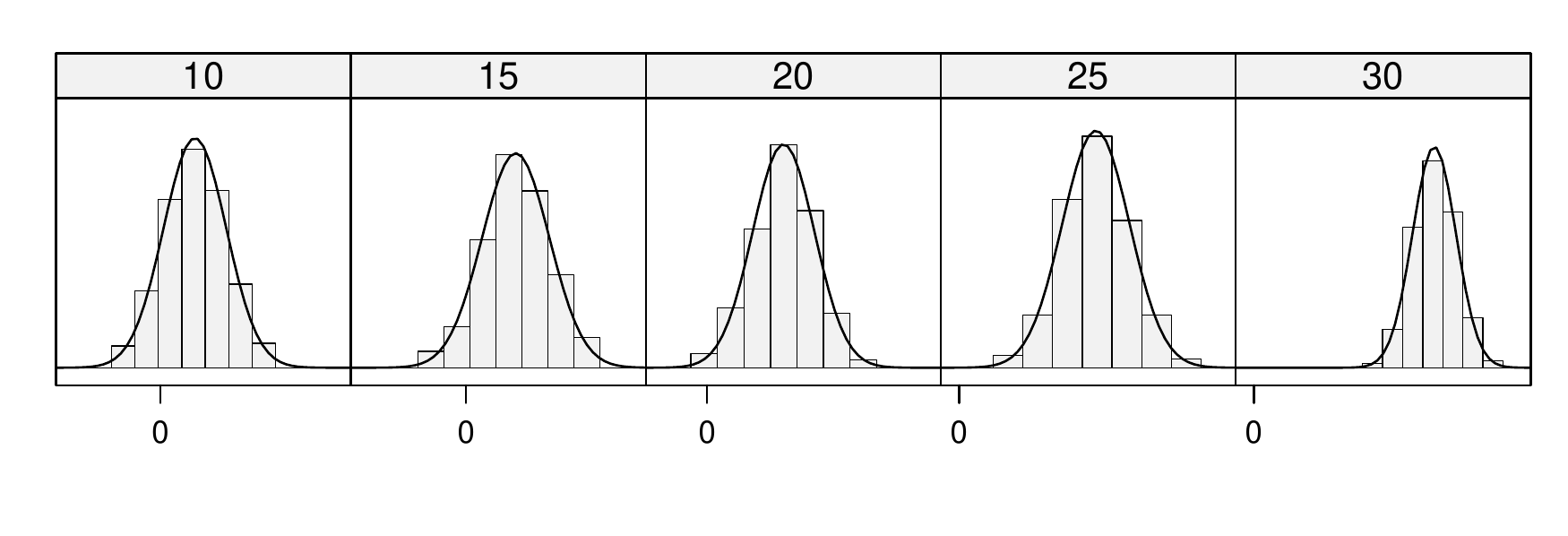}
\par\end{centering}

\vspace{-0.75cm}

\begin{centering}
\includegraphics[width=0.8\textwidth]{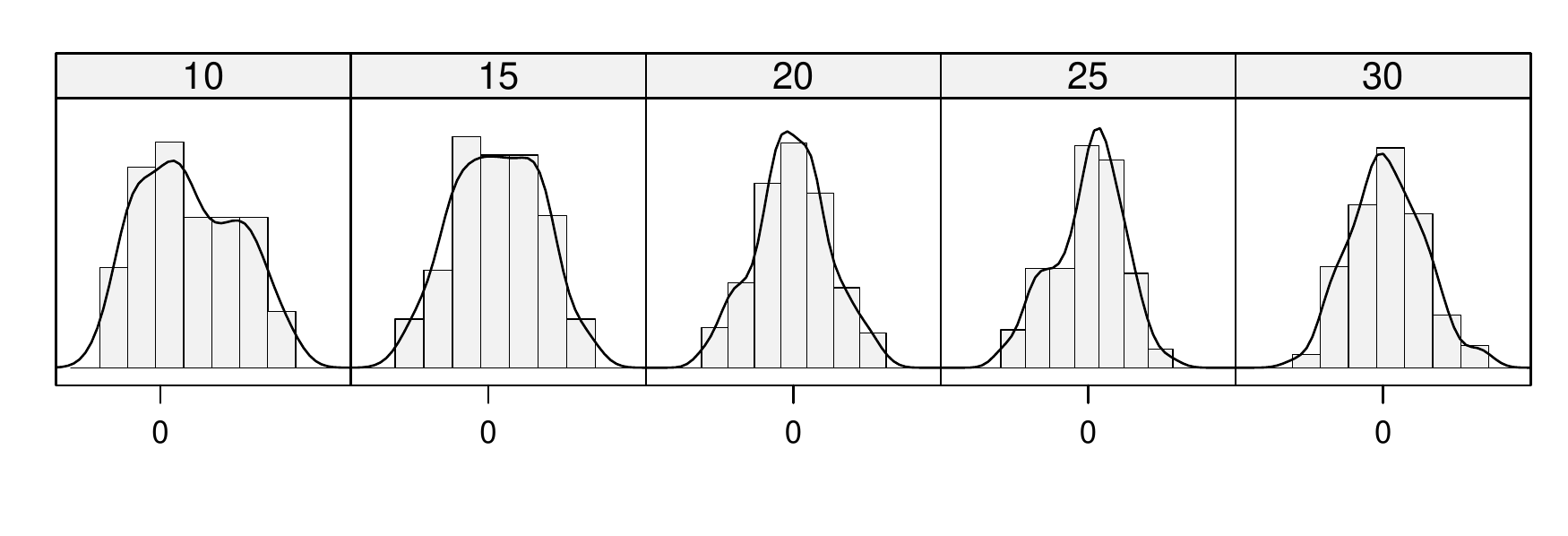}
\par\end{centering}

\centering{}\vspace{-0.75cm}
\caption{Marginal distributions of the first variable of Griewank with $\textrm{GCEDA\ensuremath{{}_{\textrm{g}}}}$
(top) and $\textrm{GCEDA\ensuremath{{}_{\textrm{e}}}}$ (bottom) in
the generations $10$, $15$, $20$, $25$ and $30$. \label{fig:griewank-margins}}
\end{figure}

\item In problems where UMDA exhibits good performance, the introduction
of correlations by GCEDA seems to be harmful.

Sphere, Griewank and Ackley can be easily optimized by UMDA as far
as the marginal information is enough for finding the global optimum.
GCEDA requires to compute many parameters and larger populations are
needed to estimate them reliably. Figure~\ref{fig:sphere-umda-gceda-boxplot}
illustrates this issue in the Sphere function. We run $\textrm{UMDA\ensuremath{{}_{\textrm{g}}}}$
with its critical population. For $\textrm{GCEDA\ensuremath{{}_{\textrm{g}}}}$
we use different population sizes, including the critical population
of these two algorithms (86 and 325, respectively). The box-plot shows
that $\textrm{GCEDA\ensuremath{{}_{\textrm{g}}}}$ achieves the means
and variances of $\textrm{UMDA\ensuremath{{}_{\textrm{g}}}}$ but
uses larger populations.

\begin{figure}[t]
\centering{}\includegraphics[width=0.85\textwidth]{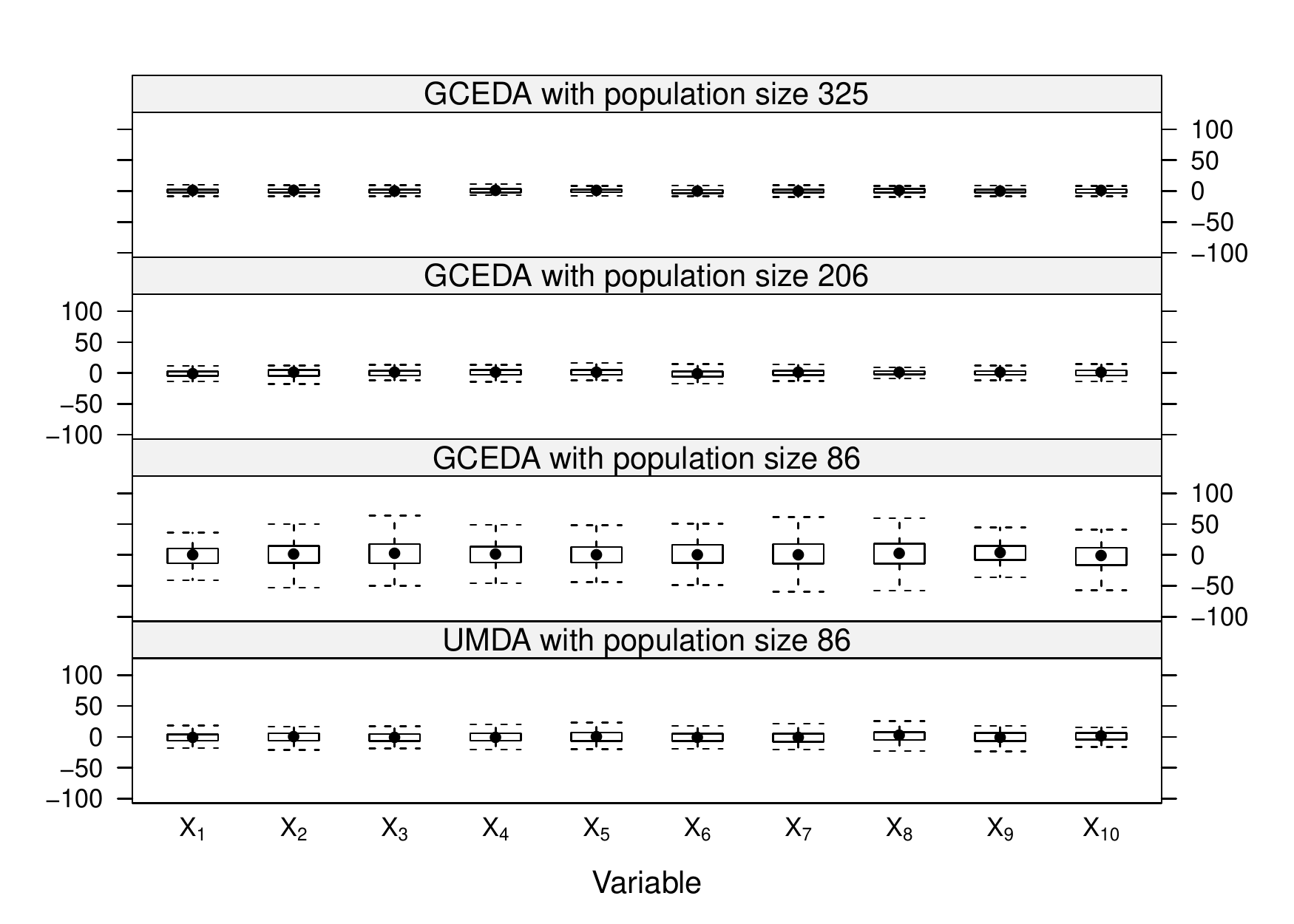}\caption{Mean and variance of each variable in the selected population at $10^{th}$
generation with $\textrm{GCEDA\ensuremath{{}_{\textrm{g}}}}$ and
$\textrm{UMDA\ensuremath{{}_{\textrm{g}}}}$ in Sphere. $\textrm{GCEDA\ensuremath{{}_{\textrm{g}}}}$
requires larger populations than $\textrm{UMDA\ensuremath{{}_{\textrm{g}}}}$.
\label{fig:sphere-umda-gceda-boxplot}}
\end{figure}

\item UMDA is not capable of optimizing Summation Cancellation.

Summation Cancellation has multivariate linear interactions between
the variables \cite{Bosman2006RealValuedEDAs}. As far as this information
is essential for finding the global optimum, UMDA fails to optimize
this function with both normal and kernel margins, while GCEDA is
successful, though this algorithm is also sensitive to the effect
of asymmetry.

\end{enumerate}
Summarizing, we can say that both aspects: the statistical properties
of the marginal distributions and the dependence structure play a
crucial role for the success of EDA optimization. In the following
sections we deal with the latter aspect.

\subsection{Effect of the Dependence Structure}

This section reports the most important results of our work. We investigate
the effect of combining different copulas, applying the truncation
strategy, and selecting the structure of C-vines and D-vines in the
performance of VEDA.

\subsubsection{Combining Different Bivariate Copulas}

In this section we assess the effect of using different types of dependences
when all the marginal distributions are normal. The experimental results
obtained with CVEDA and DVEDA in Sphere, Griewank, Ackley and Summation
Cancellation are presented in Tables \ref{tab:sphere-veda}--\ref{tab:sumcan-veda},
respectively. The studied algorithms are $\textrm{CVEDA\ensuremath{{}_{\textrm{9, greedy, g}}}}$
and $\textrm{DVEDA\ensuremath{{}_{\textrm{9, greedy, g}}}}$. The
sub-indexes mean that they perform a complete construction of the
vines ($9$ trees), use greedy heuristics to represent the stronger
dependences in the first tree, and all margins are normal.

In the investigated problems the following hold:

\begin{table}[t]
\caption{Results of VEDA in Sphere with $X_{i}\in[-600,600],\, i=1,\ldots,10$.
\label{tab:sphere-veda}}

\centering{}%
\begin{tabular}{lcccc}
\toprule 
{\small Algorithm} & {\small Success} & {\small Population} & {\small Evaluations} & {\small Best Evaluation}\tabularnewline
\midrule
{\small $\textrm{CVEDA\ensuremath{{}_{\textrm{9, greedy, g}}}}$} & {\small $30/30$} & {\small $188$} & {\small $8,033.8\pm170.5$} & {\small $6.8\textrm{E}-07\pm2.1\textrm{E}-07$}\tabularnewline
{\small $\textrm{DVEDA\ensuremath{{}_{\textrm{9, greedy, g}}}}$} & {\small $30/30$} & {\small $207$} & {\small $8,818.2\pm192.9$} & {\small $7.0\textrm{E}-07\pm1.8\textrm{E}-07$}\tabularnewline
\bottomrule
\end{tabular}
\end{table}
 
\begin{table}[t]
\caption{Results of VEDA in Griewank with $X_{i}\in[-600,600],\, i=1,\ldots,10$.
\label{tab:griewank-veda}}

\centering{}%
\begin{tabular}{lcccc}
\toprule 
{\small Algorithm} & {\small Success} & {\small Population} & {\small Evaluations} & {\small Best Evaluation}\tabularnewline
\midrule
{\small $\textrm{CVEDA\ensuremath{{}_{\textrm{9, greedy, g}}}}$} & {\small $30/30$} & {\small $213$} & {\small $9,151.9\pm452.6$} & {\small $6.5\textrm{E}-07\pm1.8\textrm{E}-07$}\tabularnewline
{\small $\textrm{DVEDA\ensuremath{{}_{\textrm{9, greedy, g}}}}$} & {\small $30/30$} & {\small $225$} & {\small $9,630.0\pm309.2$} & {\small $6.9\textrm{E}-07\pm1.5\textrm{E}-07$}\tabularnewline
\bottomrule
\end{tabular}
\end{table}
 
\begin{table}[t]
\caption{Results of VEDA in Ackley with $X{}_{i}\in[-30,30],\, i=1,\ldots,10$.
\label{tab:ackley-veda}}

\centering{}%
\begin{tabular}{lcccc}
\toprule 
{\small Algorithm} & {\small Success} & {\small Population} & {\small Evaluations} & {\small Best Evaluation}\tabularnewline
\midrule
{\small $\textrm{CVEDA\ensuremath{{}_{\textrm{9, greedy, g}}}}$} & {\small $30/30$} & {\small $213$} & {\small $11,984.8\pm184.9$} & {\small $7.9\textrm{E}-07\pm1.5\textrm{E}-07$}\tabularnewline
{\small $\textrm{DVEDA\ensuremath{{}_{\textrm{9, greedy, g}}}}$} & {\small $30/30$} & {\small $213$} & {\small $11,920.9\pm197.6$} & {\small $7.9\textrm{E}-07\pm1.3\textrm{E}-07$}\tabularnewline
\bottomrule
\end{tabular}
\end{table}
 
\begin{table}[t]
\caption{Results of VEDA in Summation Cancellation with $X_{i}\in[-0,16,\,0,16],\, i=1,\ldots,10$.
\label{tab:sumcan-veda}}

\centering{}%
\begin{tabular}{lcccc}
\toprule 
{\small Algorithm} & {\small Success} & {\small Population} & {\small Evaluations} & {\small Best Evaluation}\tabularnewline
\midrule
{\small $\textrm{CVEDA\ensuremath{{}_{\textrm{9, greedy, g}}}}$} & {\small $30/30$} & {\small $625$} & {\small $84,958.3\pm786.0$} & {\small $1.0\textrm{E}+05\pm1.1\textrm{E}-07$}\tabularnewline
{\small $\textrm{CVEDA\ensuremath{{}_{\textrm{N, 9, greedy, g}}}}$} & {\small $30/30$} & {\small $319$} & {\small $43,373.3\pm539.5$} & {\small $1.0\textrm{E}+05\pm1.3\textrm{E}-07$}\tabularnewline
{\small $\textrm{DVEDA\ensuremath{{}_{\textrm{9, greedy, g}}}}$} & {\small $30/30$} & {\small $1400$} & {\small $161,840.0\pm1,352.5$} & {\small $1.0\textrm{E}+05\pm9.3\textrm{E}-08$}\tabularnewline
{\small $\textrm{DVEDA\ensuremath{{}_{\textrm{N, 9, greedy, g}}}}$} & {\small $30/30$} & {\small $488$} & {\small $58,494.9\pm457.3$} & {\small $1.0\textrm{E}+05\pm1.3\textrm{E}-07$}\tabularnewline
\bottomrule
\end{tabular}
\end{table}

\begin{enumerate}
\item CVEDA and DVEDA exhibit a good performance in problems with both strong
and weak dependences between the variables.

\textcolor{black}{While UMDA uses the independence model and GCEDA
assumes a linear dependence structure, CVEDA and DVEDA do not assume
}the same type of dependence across all pairs of variables.\textcolor{black}{{}
The estimation procedures used by the vine-based algorithms select
among a group of candidate bivariate copulas, the one that fits the
data appropriately. CVEDA and DVEDA perform, in general, between UMDA
and GCEDA in terms of the number of function evaluations.}

\item \textcolor{black}{CVEDA exhibits better results than DVEDA in easy
problems for UMDA (Sphere, Griewank and Ackley). }

\textcolor{black}{The model used by DVEDA allows a more freely selection
of the bivariate dependences that will be explicitly modeled, while
the model used by CVEDA has a more restrictive structure. These characteristics
enable DVEDA to fit in the first tree a greater number of bivariate
copulas that represent dependences. This may explain why DVEDA requires
larger sample sizes than CVEDA, and thus more function evaluations.}

\item \textcolor{black}{CVEDA has much better results than DVEDA in Summation
Cancellation. }

\textcolor{black}{Summation Cancellation reaches its global optimum
when the sum in the denominator of the fraction is zero. The $i$-th
term of this sum is the sum of the first $i$ variables of the function.
Thus, the first variables have a greater influence in the value of
the sum. The selected populations reflect these characteristics including
stronger associations between the first variables and the next ones.
A C-vine structure provides a more appropriate modeling of this situation
than a D-vine structure, since it is possible to find a variable that
governs the interactions in the sample. However, as it was pointed
out before, here the interesting issue is the success of GCEDA. The
explanation is simple. On one hand, Summation Cancellation has multivariate
linear interactions between the variables \cite{Bosman2006RealValuedEDAs}.
On the other hand, the multivariate normal distribution is indeed,
a linear model of interactions. }

\item Combining normal and non-normal copulas worsens the results of the
vine-based algorithms\textbf{ }in Summation Cancellation.

Since the multivariate linear interactions of Summation Cancellation
are readily modeled with a multivariate normal dependence structure,
GCEDA has better performance than vine-based EDAs, which can fit copulas
of different families (Tables~\ref{tab:sumcan-umda-gceda} and \ref{tab:sumcan-veda}).
We repeated the experiments using only product and normal copulas.
The results show similar performance of $\textrm{CVEDA\ensuremath{{}_{\textrm{N, 9, greedy, g}}}}$,
$\textrm{DVEDA\ensuremath{{}_{\textrm{N, 9, greedy, g}}}}$ and GCEDA,
being CVEDA slightly better than DVEDA. 

\end{enumerate}
Regarding the results presented in this section, we can summarize
that EDAs using pair-copula constructions exhibit a more robust behavior
than EDAs using multivariate product or normal copula in the given
set of benchmark functions.

\subsubsection{Truncation of C-vines and D-vines}

In order to reduce the number of levels of the pair-copula decompositions,
and hence simplify the constructions, we apply two different approaches:
the truncation level is given as a parameter or it is determined by
a model selection procedure based on AIC or BIC (see Section \ref{sub: Estimation}).
We study the effect of both strategies in the Sphere and Summation
Cancellation functions, as examples of problems with week and strong
correlated variables. The following algorithms are compared:
\begin{itemize}
\item $\textrm{CVEDA\ensuremath{{}_{\textrm{3, greedy, g}}}}$ and $\textrm{DVEDA\ensuremath{{}_{\textrm{3, greedy, g}}}}$
truncate the vines at the third tree.
\item $\textrm{CVEDA\ensuremath{{}_{\textrm{6, greedy, g}}}}$ and $\textrm{DVEDA\ensuremath{{}_{\textrm{6, greedy, g}}}}$
truncate the vines at the sixth tree.
\item $\textrm{CVEDA\ensuremath{{}_{\textrm{AIC, greedy, g}}}}$ and $\textrm{DVEDA\ensuremath{{}_{\textrm{AIC, greedy, g}}}}$
determine the required number of trees using AIC.
\item $\textrm{CVEDA\ensuremath{{}_{\textrm{BIC, greedy, g}}}}$ and $\textrm{DVEDA\ensuremath{{}_{\textrm{BIC, greedy, g}}}}$
determine the required number of trees using BIC. 
\end{itemize}
The results of the experiments in Sphere and Summation Cancellation
are presented in Tables \ref{tab:sphere-veda-truncation} and \ref{tab:sumcan-veda-truncation},
respectively. The main results are summarized in the following points:

\begin{table}[t]
\caption{Results of VEDA with truncation in Sphere with $X_{i}\in[-600,600],\, i=1,\ldots,10$.
\label{tab:sphere-veda-truncation}}

\centering{}{\footnotesize }%
\begin{tabular}{lcccc}
\toprule 
{\small Algorithm} & {\small Success} & {\small Population} & {\small Evaluations} & {\small Best Evaluation}\tabularnewline
\midrule
{\small $\textrm{CVEDA\ensuremath{{}_{\textrm{3, greedy, g}}}}$} & {\small $30/30$} & {\small $175$} & {\small $7,536.6\pm151.9$} & {\small $6.5\textrm{E}-07\pm2.2\textrm{E}-07$}\tabularnewline
{\small $\textrm{CVEDA\ensuremath{{}_{\textrm{6, greedy, g}}}}$} & {\small $30/30$} & {\small $191$} & {\small $8,174.8\pm176.6$} & {\small $6.7\textrm{E}-07\pm1.9\textrm{E}-07$}\tabularnewline
{\small $\textrm{CVEDA\ensuremath{{}_{\textrm{AIC, greedy, g}}}}$} & {\small $30/30$} & {\small $163$} & {\small $7,106.8\pm139.3$} & {\small $6.6\textrm{E}-07\pm2.0\textrm{E}-07$}\tabularnewline
{\small $\textrm{CVEDA\ensuremath{{}_{\textrm{BIC, greedy, g}}}}$} & {\small $30/30$} & {\small $113$} & {\small $5,017.2\pm134.6$} & {\small $6.8\textrm{E}-07\pm1.6\textrm{E}-07$}\tabularnewline
{\small $\textrm{DVEDA\ensuremath{{}_{\textrm{3, greedy, g}}}}$} & {\small $30/30$} & {\small $191$} & {\small $8,149.3\pm161.2$} & {\small $6.5\textrm{E}-07\pm1.8\textrm{E}-07$}\tabularnewline
{\small $\textrm{DVEDA\ensuremath{{}_{\textrm{6, greedy, g}}}}$} & {\small $30/30$} & {\small $207$} & {\small $8,818.2\pm128.6$} & {\small $6.9\textrm{E}-07\pm1.8\textrm{E}-07$}\tabularnewline
{\small $\textrm{DVEDA\ensuremath{{}_{\textrm{AIC, greedy, g}}}}$} & {\small $30/30$} & {\small $163$} & {\small $6,992.7\pm144.2$} & {\small $6.5\textrm{E}-07\pm1.9\textrm{E}-07$}\tabularnewline
{\small $\textrm{DVEDA\ensuremath{{}_{\textrm{BIC, greedy, g}}}}$} & {\small $30/30$} & {\small $138$} & {\small $6,026.0\pm127.2$} & {\small $7.0\textrm{E}-07\pm2.2\textrm{E}-07$}\tabularnewline
\bottomrule
\end{tabular}
\end{table}
 
\begin{table}[t]
\caption{Results of VEDA with truncation in Summation Cancellation with $X_{i}\in[-0,16,\,0,16]$,
$i=1,\ldots,10$.\label{tab:sumcan-veda-truncation}}

\centering{}{\small }%
\begin{tabular}{lcccc}
\toprule 
{\small Algorithm} & {\small Success} & {\small Population} & {\small Evaluations} & {\small Best Evaluation}\tabularnewline
\midrule
{\small $\textrm{CVEDA\ensuremath{{}_{\textrm{3, greedy, g}}}}$} & {\small $0/30$} & {\small $2000$} & {\small $500,000.0\pm0.0$} & {\small $2.6\textrm{E}+03\pm3.4\textrm{E}+03$}\tabularnewline
{\small $\textrm{CVEDA\ensuremath{{}_{\textrm{6, greedy, g}}}}$} & {\small $0/30$} & {\small $2000$} & {\small $500,000.0\pm0.0$} & {\small $3.7\textrm{E}+04\pm3.2\textrm{E}+04$}\tabularnewline
{\small $\textrm{CVEDA\ensuremath{{}_{\textrm{AIC, greedy, g}}}}$} & {\small $30/30$} & {\small $650$} & {\small $90,003.3\pm1,262.8$} & {\small $1.0\textrm{E}+05\pm1.2\textrm{E}-07$}\tabularnewline
{\small $\textrm{CVEDA\ensuremath{{}_{\textrm{BIC, greedy, g}}}}$} & {\small $30/30$} & {\small $800$} & {\small $108,506.6\pm1,647.3$} & {\small $1.0\textrm{E}+05\pm9.8\textrm{E}-08$}\tabularnewline
{\small $\textrm{DVEDA\ensuremath{{}_{\textrm{3, greedy, g}}}}$} & {\small $0/30$} & {\small $2000$} & {\small $500,000.0\pm0.0$} & {\small $8.4\textrm{E}+04\pm2.5\textrm{E}+04$}\tabularnewline
{\small $\textrm{DVEDA\ensuremath{{}_{\textrm{6, greedy, g}}}}$} & {\small $10/30$} & {\small $2000$} & {\small $412,133.3\pm12,8711.1$} & {\small $9.9\textrm{E}+04\pm1.7\textrm{E}+02$}\tabularnewline
{\small $\textrm{DVEDA\ensuremath{{}_{\textrm{AIC, greedy, g}}}}$} & {\small $30/30$} & {\small $1300$} & {\small $152,750.0\pm1,404.1$} & {\small $1.0\textrm{E}+05\pm1.0\textrm{E}-07$}\tabularnewline
{\small $\textrm{DVEDA\ensuremath{{}_{\textrm{BIC, greedy, g}}}}$} & {\small $26/30$} & {\small $2000$} & {\small $285,000.0\pm100,221.0$} & {\small $9.9\textrm{E}+04\pm6.9\textrm{E}-03$}\tabularnewline
\bottomrule
\end{tabular}
\end{table}

\begin{enumerate}
\item The algorithms that use the truncation strategy based on AIC or BIC
exhibit a more robust behavior.

The necessary number of trees depends on the characteristics of the
function being optimized. In the Sphere function, a small number of
trees is quite enough, while in Summation Cancellation it is preferable
to expand the pair-copula decomposition completely. In both functions
the better results are obtained when the truncation level is determined
by a model selection procedure based on AIC or BIC, since cutting
the model arbitrarily could cause that important dependences are not
represented. The latter was the strategy applied in \cite{Salinas-Gutierrez2010DvineEDA},
where a D-vine with normal copulas was only expanded up to the second
tree. A combination of both strategies could be an appropriate solution.

\item For VEDA the truncation method based on AIC is preferable than the
truncation based on BIC.

In the Sphere function, the vine-based EDAs that use truncation based
on BIC perform better than those based on AIC. The opposite occurs
in Summation Cancellation, where $\textrm{DVEDA\ensuremath{{}_{\textrm{BIC, greedy, g}}}}$
fail in the $30$ runs. Both situations are caused by the term that
penalizes the number of parameters in these metrics. BIC prefers models
with less number of copulas than AIC \cite{Brechmann2010DiplomaThesis},
which is good for Sphere, but compromises the convergence of the algorithms
in Summation Cancellation. The algorithms using AIC have a good performance
in both functions. Specifically, in Sphere the number of trees was
never greater than three with CVEDA and four with DVEDA; in Summation
Cancellation both algorithms perform complete construction of the
vines (nine trees).

\end{enumerate}
In the following section, we study the importance of the selection
of the bivariate dependences explicitly modeled in the first tree
of C-vines and D-vines.

\subsubsection{Selection of the Structure of C-vines and D-vines}

The aim of this section is to assess the importance of selecting an
appropriate ordering of the variables in the pair-copula decomposition
for the optimization with vine-based EDAs. 

Here we repeat the experiments with Sphere and Summation Cancellation,
but this time the variables in the first tree in the decomposition
are ordered randomly instead of representing the strongest bivariate
dependences.  The instances of the algorithms selected in these experiments
are those that showed the best performance in the truncation experiments
of the previous section. The results are presented in Tables \ref{tab:sphere-veda-structure}
and \ref{tab:sumcan-veda-structure}.

\begin{table}[t]
\caption{Results of VEDA with a random selection of the structure of the first
tree of the vines at each generation in Sphere with $X{}_{i}\in[-600,600],\, i=1,\ldots,10$.
\label{tab:sphere-veda-structure}}

\centering{}%
\begin{tabular}{lcccc}
\toprule 
{\small Algorithm} & {\small Success} & {\small Population} & {\small Evaluations} & {\small Best Evaluation}\tabularnewline
\midrule
{\small $\textrm{CVEDA\ensuremath{{}_{\textrm{BIC, random, g}}}}$} & {\small $30/30$} & {\small $100$} & {\small $4,523.3\pm100.6$} & {\small $6.9\textrm{E}-07\pm1.8\textrm{E}-07$}\tabularnewline
{\small $\textrm{DVEDA\ensuremath{{}_{\textrm{BIC, random, g}}}}$} & {\small $30/30$} & {\small $100$} & {\small $4,526.6\pm114.2$} & {\small $6.6\textrm{E}-07\pm1.6\textrm{E}-07$}\tabularnewline
\bottomrule
\end{tabular}
\end{table}
 
\begin{table}[t]
\caption{Results of VEDA with a random selection of the structure of the first
tree of the vines at each generation in Summation Cancellation with
$X_{i}\in[-0,16,\,0,16],\, i=1,\ldots,10$. \label{tab:sumcan-veda-structure}}

\centering{}%
\begin{tabular}{lcccc}
\toprule 
{\small Algorithm} & {\small Success} & {\small Population} & {\small Evaluations} & {\small Best Evaluation}\tabularnewline
\midrule
{\small $\textrm{CVEDA\ensuremath{{}_{\textrm{AIC, random, g}}}}$} & {\small $30/30$} & {\small $775$} & {\small $110,360.0\pm2,020.9$} & {\small $1.0\textrm{E}+05\pm1.1\textrm{E}-07$}\tabularnewline
{\small $\textrm{DVEDA\ensuremath{{}_{\textrm{AIC, random, g}}}}$} & {\small $30/30$} & {\small $1500$} & {\small $255,900.0\pm5,205.7$} & {\small $1.0\textrm{E}+05\pm1.2\textrm{E}-07$}\tabularnewline
\bottomrule
\end{tabular}
\end{table}

In the Sphere function, the algorithms that use a random structure
exhibit a better performance, since the number of product copulas
that are fitted is greater. In this case, the estimated model resembles
independence model used by UMDA, which indeed exhibits the best performance
with the Sphere function. The opposite occurs with Summation Cancellation,
where the use of a random structure in the first tree causes that
important correlations for an efficient search are not represented,
which deteriorates the performance of the algorithms in terms of the
number of function evaluations. The main conclusion of this part is
that it is necessary to make a careful selection of the structure
of the pair-copula decomposition. The representation of the strongest
dependences is important in order to construct more robust vine-based
EDAs.

\section{Conclusions\label{sec:conclusions}}

This paper introduces a class of EDAs called VEDAs. Two algorithms
of this class are presented: CVEDA and DVEDA, which model the search
distributions using C-vines and D-vines, respectively.

The copula EDAs based on vines are more flexible than those based
on the multivariate product and normal copulas, because the PCC models
can describe a richer variety of dependence patterns. Our empirical
investigation confirms the robustness of CVEDA and DVEDA in both strong
and weak correlated problems.

We have found that building the complete structure of the vine is
not always necessary. However, cutting the model at a tree selected
arbitrarily could cause that important dependences are not represented.
A more appropriate global strategy could be to combine setting a maximum
number of trees with a model selection technique, such as the truncation
method based on AIC or BIC. We also found that it is important to
make a conscious selection of the pairwise dependences represented
explicitly in the model.

Our findings show that both the statistical properties of the margins
and the dependence structure play a crucial role in the success of
optimization. The use of copulas and vines in EDAs represents a new
way to deal with more flexible search distributions and different
sources of complexity that arise in optimization.

As future research we consider to extend the class of VEDAs with regular
vines. Our algorithms have been used in the optimization of test functions,
such as the ones proposed in CEC 2005 benchmark \cite{Suganthan2005CEC2005Benchmark}.
In general, these functions display independence or linear correlations.
In the future, we will seek problems with relevant dependences to
the vine models studied in this work.

\appendix

\section{Expressions of the $h$ and $h^{-1}$ Functions of Various Bivariate
Copulas\label{sec:h-functions}}

The pair-copulas used in this work are product, normal, Student's
$t$, Clayton, rotated Clayton, Gumbel and rotated Gumbel. This appendix
contains the definition of these copulas and the $h$ and $h^{-1}$
functions required to use this copulas in pair-copula constructions.

\subsection*{The Bivariate Product Copula}

An immediate consequence of Sklar's theorem is that two random variables
are independent if and only if their underlying copula is $C_{\textrm{I}}\left(u,v\right)=uv$.
For this copula $h_{\textrm{I}}(x,v)=x$ and $h_{\textrm{I}}^{-1}(u,v)=u$.

\subsection*{The Bivariate Normal Copula}

The distribution function of the bivariate normal copula is given
by

\[
C_{\textrm{N}}(u,v;\rho)=\Phi_{\rho}(\Phi^{-1}(u),\Phi^{-1}(v)),
\]
where $\Phi_{\rho}$ is the bivariate normal distribution function
with correlation parameter~$\rho$ and $\Phi^{-1}$ is the inverse
of the standard univariate normal distribution function. For this
copula the $h$ and $h^{-1}$ functions are 

\[
h_{\textrm{N}}\left(x,v;\rho\right)=\Phi\left(\frac{\Phi^{-1}\left(x\right)-\rho\,\Phi^{-1}\left(v\right)}{\sqrt{1-\rho^{2}}}\right),
\]
\[
h_{\textrm{N}}^{-1}\left(u,v;\rho\right)=\Phi\left(\Phi^{-1}\left(u\right)\sqrt{1-\rho^{2}}+\rho\,\Phi^{-1}\left(v\right)\right).
\]

\noindent The derivation of these formulas are given in \cite{Aas2009PairCopulaConstructions}.

\subsection*{The Bivariate Student's $t$ Copula}

The distribution function of the bivariate Student's $t$ copula is
given by

\[
C_{\textrm{t}}(u,v;\rho,\nu)=t_{\rho,\nu}(t_{\nu}^{-1}(u),t_{\nu}^{-1}(v)),
\]
where $t_{\rho,\nu}$ is the distribution function of the bivariate
Student's $t$ distribution with correlation parameter $\rho$ and
$\nu$ degrees of freedom and $t_{\nu}^{-1}$ is the inverse of the
univariate Student's $t$ distribution function with $\nu$ degrees
of freedom. For this copula the $h$ and $h^{-1}$ functions are 

\[
h_{\textrm{t}}\left(x,v;\rho,\nu\right)=t_{\nu+1}\left(\frac{t_{\nu}^{-1}\left(x\right)-\rho\, t_{\nu}^{-1}\left(v\right)}{\sqrt{\frac{\left(\nu+\left(t_{\nu}^{-1}\left(v\right)\right)^{2}\right)\left(1-\rho^{2}\right)}{\nu+1}}}\right),
\]
\[
h_{\textrm{t}}^{-1}\left(u,v;\rho,\nu\right)=t_{v}\left(t_{v+1}^{-1}\left(u\right)\sqrt{\frac{\left(\nu+\left(t_{\nu}^{-1}\left(v\right)\right)^{2}\right)\left(1-\rho^{2}\right)}{\nu+1}}+\rho\, t_{\nu}^{-1}\left(v\right)\right).
\]

\noindent The derivation of these formulas are given in \cite{Aas2009PairCopulaConstructions}.

\subsection*{The Bivariate Clayton Copula}

The distribution function of the bivariate Clayton copula is given
by

\begin{equation}
C_{\textrm{C}}\left(u,v;\theta\right)=\left(u^{-\theta}+v^{-\theta}-1\right)^{-1/\theta},\label{eq:clayton-copula}
\end{equation}
where $\theta>0$ is a parameter controlling the dependence. Perfect
dependence is obtained when $\theta\rightarrow\infty$, while $\theta\rightarrow0$
implies independence. For this copula the $h$ and $h^{-1}$ functions
are

\[
h_{\textrm{C}}\left(x,v;\theta\right)=v^{-\theta-1}\left(x^{-\theta}+v^{-\theta}-1\right)^{-1-1/\theta},
\]
\[
h_{\textrm{C}}^{-1}\left(u,v;\theta\right)=\left(\left(uv^{\theta+1}\right)^{-\theta/(\theta+1)}+1-v^{-\theta}\right)^{-1/\theta}.
\]

\noindent The derivation of these formulas are given in \cite{Aas2009PairCopulaConstructions}.

\subsection*{The Bivariate Rotated Clayton Copula}

The bivariate Clayton copula, as defined in (\ref{eq:clayton-copula}),
can only capture positive dependence. Following the transformation
used in \cite{Brechmann2010DiplomaThesis}, we consider a $90$ degrees
rotated version of this copula. The distribution function of the bivariate
rotated Clayton copula is obtained as

\[
C_{\textrm{RC}}(u,v;\theta)=u-C_{\textrm{C}}(u,1-v;-\theta),
\]
where $\theta<0$ is a parameter controlling the dependence and $C_{\textrm{C}}$
denotes the distribution function of the bivariate Clayton copula.
For this copula the $h$ and $h^{-1}$ functions are

\[
h_{\textrm{RC}}(x,v;\theta)=h_{\textrm{C}}(x,1-v;-\theta)
\]
 and 
\[
h_{\textrm{RC}}^{-1}(u,v;\theta)=h_{\textrm{C}}^{-1}(u,1-v;-\theta),
\]
where $h_{\textrm{C}}$ and $h_{\textrm{C}}^{-1}$ denote the expressions
of the $h$ and $h^{-1}$ functions for the bivariate Clayton copula.

\subsection*{The Bivariate Gumbel Copula}

The distribution function of a bivariate Gumbel copula is given by

\[
C_{\textrm{G}}(u,v;\theta)=\exp\left(-\left(\left(-\log\, u\right)^{\theta}+\left(-\log\, v\right)^{\theta}\right)^{1/\theta}\right),
\]
where $\theta\geq1$ is a parameter controlling the dependence. Perfect
dependence is obtained when $\theta\rightarrow\infty$, while $\theta=1$
implies independence. The $h$ function is 
\[
h_{G}\left(x,v;\theta\right)=C_{G}\left(x,v;\theta\right)\frac{1}{v}\left(-\log\, v\right)^{\theta-1}\left[\left(-\log\, x\right)^{\theta}+\left(-\log\, v\right)^{\theta}\right]^{1/\theta-1},
\]
but $h_{\textrm{G}}^{-1}$ cannot be written in closed form; therefore,
we obtain it numerically using Brent's method \cite{Brent1973MinimizationWithoutDerivatives}.
The derivation of these formulas are given in \cite{Aas2009PairCopulaConstructions}.

\subsection*{The Bivariate Rotated Gumbel Copula}

The bivariate Gumbel copula can only represent positive dependence.
As for the bivariate Clayton copula and following the transformation
used in \cite{Brechmann2010DiplomaThesis}, we also consider a $90$
degrees rotated version of the bivariate Gumbel copula. The distribution
function of the bivariate rotated Gumbel copula is defined as

\[
C_{\textrm{RG}}(u,v;\theta)=u-C_{G}(u,1-v;-\theta),
\]
where $\theta<-1$ is a parameter controlling the dependence and $C_{\textrm{G}}$
denotes the distribution function of the bivariate Gumbel copula.
For this copula the $h$ and $h^{-1}$ functions are

\[
h_{\textrm{RG}}(x,v;\theta)=h_{\textrm{G}}(x,1-v;-\theta)
\]
 and

\[
h_{\textrm{RG}}^{-1}(u,v;\theta)=h_{\textrm{G}}^{-1}(u,1-v;-\theta),
\]
where $h_{\textrm{G}}$ and $h_{\textrm{G}}^{-1}$ denote the expressions
of the $h$ and $h^{-1}$ functions for the bivariate Gumbel copula.

\end{document}